\documentclass[10pt,twocolumn]{article}

\usepackage[margin=0.75in]{geometry}
\usepackage{times}
\usepackage{mathptmx}
\usepackage{amsmath, amssymb}
\usepackage{graphicx}
\usepackage{booktabs}
\usepackage{multirow}
\usepackage{hyperref}
\usepackage{url}
\usepackage{caption}
\usepackage{subcaption}
\usepackage{algorithm}   
\usepackage{amsthm}
\usepackage[numbers]{natbib}
\usepackage{bm}

\usepackage{algorithm}
\usepackage{algpseudocode}

\newtheorem{proposition}{Proposition}
\newcommand{\NE}{N_E}
\newcommand{\Wtern}{\hat{W}_{\text{tern}}}
\newcommand{\BF}{\mathcal{B}}

\title{\bf ButterflyViT: 354$\times$ Expert Compression for Edge Vision Transformers}
\author{
Aryan Karmore \\
Indian Institute of Information Technology, Nagpur \\
\texttt{bt24csd009@iiitn.ac.in}
}
\date{}

\begin{document}

\twocolumn

\maketitle
\begin{abstract}
Deploying sparse Mixture of Experts(MoE) Vision Transformers remains a challenge due to linear expert memory scaling. Linear memory scaling stores $N$ independent expert weight matrices requiring $\mathcal{O}(N_E \cdot d^2)$ memory, which exceeds edge devices memory budget. Current compression methods like quantization, pruning and low-rank factorization reduce constant factors but leave the scaling bottleneck unresolved. We introduce ButterflyViT, a method that treats experts not as independent weight matrices but as geometric reorientations of a unified shared quantized substrate. Diversity among experts arises from viewing different angles of shared capacity, not from redundant storage.
By applying learned rotations to a shared ternary prototype, each expert yields $\mathcal{O}(d_{\text{model}} \cdot d_{\text{ff}} + N_E \cdot n_\ell \cdot d)$ memory which is sub-linear in the number of experts. To address the unique challenges of vision, a spatial smoothness regulariser is introduced that penalises routing irregularities between adjacent patch tokens, turning patch correlation into a training signal.
Across image classification tasks on CIFAR-100, ButterflyViT achieves 354$\times$ memory reduction at 64 experts with negligible accuracy loss. ButterflyViT allows multiple experts to fit on edge-constrained devices showing that geometric parameterization breaks linear scaling.
\end{abstract}
\section{Introduction}

A (Mixture of Expert) MoE ViT layer with 64 experts having dimension $d=256$ requires 939 MB of memory, this exceeds edge device limits. This is the result of an architectural assumption that each expert requires separate parameters.

Current compression methods retain linear memory scaling. Quantization based approaches reduce bit-width and achieve substantial compression but retains $O(N \cdot d^2)$ memory growth. Even 2 bit quantization requires hundreds of MB per layer at 64 experts. The important question is whether whether $N$ experts need $O(N)$ separate parameter sets or not?

In light of recent research on linear model connection, we propose that rather than being stored separately, experts can be formed from a shared prototype through straightforward transformations. This arrangement introduces sub-linear memory scaling.

ButterflyViT parameterizes $N$ experts as learned rotations of a single ternary-quantized substrate. Each expert $W_i$ is defined as:
\[
W_i \;\approx\; \mathcal{B}(\phi_i) \cdot W_{\text{base}}
\cdot \mathcal{B}(\theta_i)^\top
\]
where $W_{\text{base}} \in \{-1, 0, +1\}^{d_{\text{ff}} \times d_{\text{model}}}$
is a shared 1.58-bit weight matrix and $\mathcal{B}(\phi_i), \mathcal{B}(\theta_i)$
are expert-specific butterfly matrices with $O(n_\ell \cdot d)$ parameters. Experts are never explicitly materialized. Inference proceeds by applying a rotation and a ternary matrix multiply. This yields $\mathcal{O}(d_{\text{model}} \cdot d_{\text{ff}} + N_E \cdot n_\ell \cdot d)$ memory.

This structural change has two benefits:(i) Memory compression improves with expert count achieving a 354x compression at 64 experts, (ii) Per-expert input rotations suppress activation outliers.

ButterflyViT matches StandardMoE accuracy on CIFAR-100 image classification
while enabling deployments previously unattainable . At 64 experts ButterflyViT uses 0.379 MB of memory, compared to 939 MB for the baseline, and trains stably without recovery stages.

\textbf{Contributions}
\begin{itemize}
    \item We introduce ButterflyViT, combining ternary quantization with learned Butterfly rotations to achieve $\mathcal{O}(d_{\text{model}} \cdot d_{\text{ff}} + N_E \cdot n_\ell \cdot d)$ memory complexity.
    \item We demonstrate 354$\times$ compression at 64 experts with competitive accuracy, enabling massively parallel edge deployment and up to 99.5\% memory bandwidth energy reduction under standard DRAM energy models.
    
\end{itemize}

\section{Literature Review}
This paper is inspired from ButterflyMoE\cite{karmore2026butterflymoe} which introduced orbital parameterization of MoE experts using butterfly rotations of a shared substrate in language models showing sub linear memory scaling.
This work extends this to vision transformers, introducing spatial smoothness regularization as a vision specific contribution and providing the first benchmark results on image classification.

Prior Work is organized across three axes:
\begin{itemize} 
    \item ViT architectures
    \item Model Compression and Quantization
    \item Sparse MoE for ViT
\end{itemize}

\subsection{ViT architectures}
Vision Transformer \cite{dosovitskiy2020image} showed that patch based token models perform well but suffer from quadratic attention complexity as the sequence length increases. Prior work on ViT architectures used to focus on linearising or reducing the quadratic cost of self attention. Efficent ViT research has converged on : architecture design, pruning, knowledge distillation and quantization.
Pyramid Vision Transformer \cite{wang2021pyramid} established hierarchical feature maps without convolutions. This enables dense prediction tasks with a reduced cost.
DeiT \cite{touvron2021training} showed that ViTs can be trained on ImageNet via knowledge distillation. This removed the dependency on massive pretraining datasets.
By confining attention to shifted local windows, Swin Transformer \cite{liu2021swin} addressed the quadratic complexity. This approach achieved linear complexity with respect to image size.
Orthogonal Transformer \cite{fei2022vit} uses token orthogonalisation within local groups to reduce attention cost. EfficentViT \cite{liu2023efficientvit} replaced standard multi head self attention with lightweight group attention.
All of these approaches operate on the attention module and leave the FFN layer untouched. The FFN layer is the primary target for ButterflyViT

\subsection{Model Compression and Quantization}
Post training quantization for ViTs due to the gaussian and heavy tailed activation distributions which are produced by softmax attention. ViT's with low bits are unexplored and experience performance drops compared to models which are bigger with information distortion in the attention map as the bottleneck. Using mixed precision weights Liu et al. \cite{liu2021post} tackles post quantization in ViTs. PTQ4ViT \cite{yuan2022ptq4vit} uses twin uniform quantiztion combined with a Hessian guided metric for scaling factor selection. This method produces good results at 8 bit quantization on ImageNet.
To fully enable the quantized inference FQ-ViT\cite{lin2021fq} showed power of two factor quantization and log interger based softmax.
Q-ViT\cite{li2022q} proposed an information rectification module which targets the attention mechanism, achieving 6.14× theoretical speedup on ViT-S while surpassing full-precision accuracy by 1\% on ImageNet. 
I-ViT\cite{li2023vit} used integer only arithematic for the full computational graph which achieves substantial latency gains on Swin and DeiT architectures.
All of these works target the entire network uniformly and apply quantization post-hoc to pretrained models. ButterflyViT changes this as ternary quantization is applied selectively to shared susbtrate and the quantization is learned end to end using STE rather than applied post training\cite{yin2019understanding}

\subsection{Sparse MoE for ViT}
V-MoE\cite{riquelme2021scaling} established sparse MoE for vision by replacing the dense FFN layers of ViT by sparsely gated independent expert FFNs, this method achieved substantial accuracy gains and required as little as half the compute at inference time. V-MoE also introduced Batch Priority Routing which uses expert capacity constraints to skip uniformative patches.
 M3ViT \cite{fan2022m3vit} showed sparse MoE ViT to multi-task learning replacing dense FFN layers with task specific sparsely activated MoE achieving 9.23× lower latency and energy on FPGA compared to a cache-based MoE ViT baseline.

Mobile V-MoE\cite{daxberger2023mobile} showed a scaled down version of sparse MoE routing entire images rather than individual patches to experts using class guided routing outperforming dense ViT.

 Across these works, MoE is used to increase the model capacity instead of reducing the parameter footprint. ButterflyViT is the first work to invert this objective by using a shared ternary substrate and per expert butterfly rotations\cite{dao2019learning} to achieve sub linear expert memory scaling. Butterfly Matrices \cite{dao2019learning} parameterize orthogonal transformations with $\mathcal{O}(d \log d)$ parameters by using block-diagonal Givens rotations, enabling fast $\mathcal{O}(d \log d)$ matrix-vector products. The main objective of this work is targetting deployment efficency rather than capacity scaling,

\begin{figure}[H]
    \centering
    \includegraphics[width=0.4\textwidth]{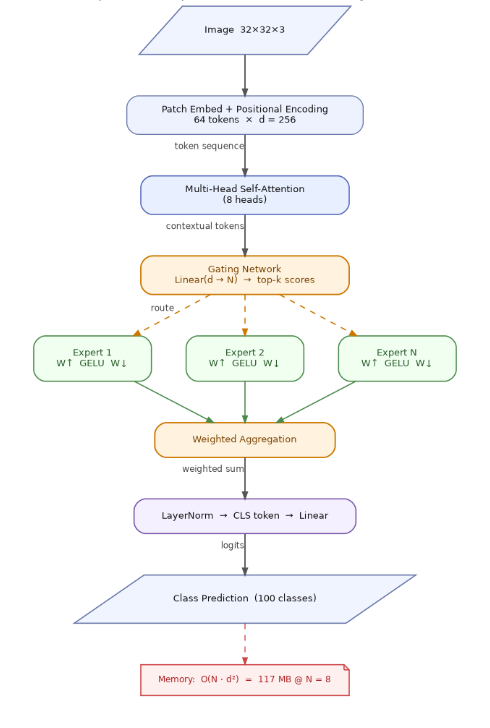}
    \caption{Standard ViT-MoE architecture}
    \label{fig:your_label}
\end{figure}
The missing primitive: a parameterization that represents $N$ experts as group-theoretic variations of a shared substrate, breaking the assumption that experts must be independently stored.

\subsection{ButterflyViT}
ButterflyViT introduces a structural insight that experts in ViT-MoE models can be parameterized as orbital variations of a single quantized prototype under learned transformations. 
Expert storage is replaced with group-orbit parameterization. This insight eliminates the need to materialize expert weights.
Motivated by the work on linear mode connectivity \cite{garipov2018loss,ainsworth2022git} which show that distinct neural networks trained on similar data occupy low dimensional manifolds connected by smooth transformations. This observation is extended to ViT-MoE based architectures by parameterizing $N$ independent matrices $\{W_i\}$ and represented as :

\begin{equation*}
W_i = \mathcal{B}(\phi_i) \cdot W_{\text{base}} \cdot \mathcal{B}(\theta_i)^T
\end{equation*}

where  $W_{\text{base}} \in \{-1, 0, +1\}^{d_{\text{ff}} \times d_{\text{model}}}$  is ternary-quantized (1.58 bits/weight) and $\mathcal{B}(\phi_i), \mathcal{B}(\theta_i)$ are expert-specific Butterfly matrices.

This achieves three properties simultaneously:

\begin{itemize}
    \item \textbf{Sub-linear memory scaling:} $\mathcal{O}(d_{\text{model}} \cdot d_{\text{ff}} + N_E \cdot n_\ell \cdot d)$ versus $\mathcal{O}(N_E \cdot d_{\text{model}} \cdot d_{\text{ff}})$
    \item \textbf{Dynamic outlier suppression:} Learned per-expert input rotations redistribute activation energy across dimensions during training, reducing quantization sensitivity.
    \item \textbf{Joint optimization:}  The resulting parameterization naturally induces expert diversity while maintaining a shared quantization-friendly substrate
\end{itemize}

The result: $354\times$ compression at 64 experts compared to standard MoE, enabling deployment scenarios that are not feasible under prior methods given current memory constraints. Unlike quantization only approaches, our compression ratio improves as $N$ increases.

\section{Methodology}
\subsection{Problem Setup}
This setup targets vision modelling on edge devices where memory is constrained.
Let $\mathcal{D} = \{(x^{(i)}, y^{(i)})\}_{i=1}^N$ denote a vision training 
dataset where $x^{(i)} \in \mathbb{R}^{3 \times H \times W}$ is an RGB image(3 channels) 
and $y^{(i)} \in \{1, \ldots, C\}$ is a class label.

A standard Mixture of Experts (MoE)-ViT layer has $N_E$ independent expert FFNs. Let 
$g : \mathbb{R}^{d_{\text{model}}} \to \Delta^{N_E - 1}$ be a gating network 
that routes each patch token to the top-$k$ experts. The MoE output is:
\begin{equation}
y = \sum_{i \in \text{TopK}(g(x))} g_i(x) \cdot W_i x
\end{equation}
where $W_i \in \mathbb{R}^{d_{\text{ff}} \times d_{\text{model}}}$ denotes 
the parameters of expert $i$.

The resulting memory footprint is:
\begin{equation}
M_{\text{MoE}} = N_E \cdot d_{\text{ff}} \cdot d_{\text{model}} 
\cdot b_{\text{precision}} \;\text{bytes}
\end{equation}
where $b_{\text{precision}} = 4$ bytes (FP32). . For $N_E = 8$ experts with 
$d_{\text{model}} = 256$ and $d_{\text{ff}} = 1024$, this yields 
$M_{\text{MoE}} \approx 117$ MB, growing to 940 MB at 64 experts. This is not feasible for edge devices having a memory constraint.
The core objective is to find a method that satisfies:
\begin{itemize}
    \item Sub-linear memory scaling in $N_E$
    \item Expert diversity for visual specialisation
    \item Accuracy preservation on image classification
\end{itemize}

\subsection{Why Standard MoE fails on Edge Devices}
Standard MoE assume experts are independently parameterized with each \(W_i\) stored explicitly. This results in linear memory scaling \(O(N_E \cdot d^2)\) which leads to two critical failures:

\begin{itemize}
    \item \textbf{F1. Memory wall:}
    An 8-expert MoE with \(d = 256\) occupies 117 MB in FP32. Scaling to 64 experts, is needed for fine-grained visual specialization across object categories, textures, scene types. This requires 940 MB, which exceeds the capacity of most edge devices.

    \item \textbf{F2. Bandwidth bottleneck:}
    Even when the experts fit in DRAM, inference is dominated by memory bandwidth and energy costs due to repeated loading of \(N_E\) weight matrices from memory. Using a DRAM access energy of 6.4 pJ/bit \cite{6757323}, a 940 MB MoE consumes \(\sim\)13 mJ per forward pass, unsuitable for battery powered devices. 
\end{itemize}

\subsection{ Core Insight: Experts as Orbits of a Quantized Prototype}
A standard MoE architecture stores $N_E$ independent weight matrices
$\{W_i\}_{i=1}^{N_E}$, each of size $d_{\text{ff}} \times d_{\text{model}}$ which results in linear memory growth where memory grows linearly as we increase the number of experts. The key point of ButterflyViT is that this redundancy is not needed as expert specialisation in vision does not need independent parameters. It needs independent perspectives on shared parameters.

Rather than
learning each $W_i$ from scratch, the entire expert family is represented as a rotated view of a shared ternary prototype:
\begin{equation}
    W_i \;\approx\; \mathcal{B}(\phi_i) \; W_{\text{base}} \; \mathcal{B}(\theta_i)^\top
\end{equation}
where $W_{\text{base}} \in \{-1, 0, +1\}^{d_{\text{ff}} \times d_{\text{model}}}$ 
is stored at 1.58 bits per weight and scaled by $\gamma$ at runtime, and 
$\mathcal{B}(\theta_i),\, \mathcal{B}(\phi_i)$ are per-expert butterfly 
rotation matrices. 
$W_i$ is never materialised as these rotations are applied sequentially at runtime  which preserves the sub-linear memory footprint.

This design delivers three properties simultaneously:

\begin{itemize}
    \item \textbf{Shared representational capacity}
    $W_{\text{base}}$ captures low level visual features like color gradients, textures and edges which are useful across all token types and image regions. The memory cost becomes fixed and ammortised across all $N_E$ experts by quantizing it to 1.58 bits per weight.

    \item \textbf{Expert diversity through rotation}
    Each expert applies learnable rotations $\mathcal{B}(\theta_i)$ and $\mathcal{B}(\phi_i)$. These tokens project tokens into different subspaces of the shared substrate, allowing experts to specialise in different visual concepts like textures, background regions, object parts and edges etc depending on which orientation of $W_{\text{base}}$ best represents the tokens routed to them.

    \item \textbf{Outlier suppression} 
    Transformer activations exhibit extreme outliers which degrade ternary quantization \cite{xiao2023smoothquant}. The input rotations in ButterflyViT are trained end to end under the classification loss, learn to redistribute activations across dimensions which minimises quantization induced projection errors and suppresses outliers.

\end{itemize}

The substrate stores what to compute while the rotations determine how to orient.
The result is a MoE architecture whose expert memory scales as
$O(d_{\text{model}} \cdot d_{\text{ff}}) + N_E \cdot O(n_\ell \cdot d)$
rather than $N_E \cdot O(d_{\text{model}} \cdot d_{\text{ff}})$,
achieving compression ratios of $181\times$ at $N_E = 8$ and $354\times$
at $N_E = 64$ relative to Standard MoE.

\begin{figure}[H]
    \centering
    \includegraphics[width=0.5\textwidth]{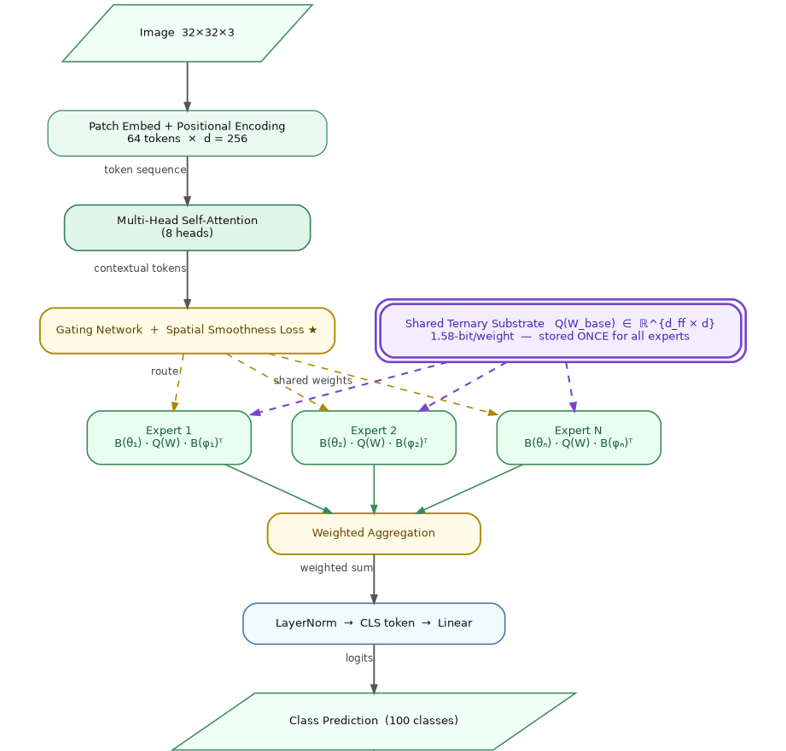}
    \caption{Top-$k$ gating instantiates experts via lightweight rotations of a shared ternary base matrix $\mathbf{W}_{\text{base}}$ .}
    \label{fig:your_label}
\end{figure}

\subsection{Parameterization via Butterfly Matrices}
Orthogonal transformations $\mathcal{R} \in O(d)$ require $O(d^2)$ parameters and therefore offer no savings in memory footprint.
This is solved using butterfly matrices \cite{dao2019learning} where rotations are parameterized with only $O(d \log d)$ scalars through a recursive block diagonal decomposition.
This is defined below:

A Butterfly matrix $\mathcal{B}(\theta) \in \mathbb{R}^{d \times d}$ with $d = 2^m$ is:
\begin{equation}
    \mathcal{B}(\theta) = \prod_{\ell=1}^{m} \mathcal{D}_\ell(\theta) \mathcal{P}_\ell
\end{equation}
where:
\begin{itemize}
    \item $\mathcal{D}_\ell(\theta)$ is block-diagonal with $2^\ell \times 2^\ell$ blocks, each applying a 2D Givens rotation:
    \begin{equation}
        \begin{bmatrix}
            \cos \alpha & -\sin \alpha \\
            \sin \alpha & \cos \alpha
        \end{bmatrix}
    \end{equation}
    \item $\mathcal{P}_\ell$ is a fixed permutation (perfect shuffle).
    \item $\theta = \{\alpha_{\ell,j}\}_{\ell=1,\ldots,m;\; j=1,\ldots,d/2}$ are learnable angles.
\end{itemize}

\subsubsection{Complexity}
\begin{itemize}
    \item \textbf{Storage:} $O(d \log d)$ parameters in the full theoretical construction. $n_\ell = 2$ butterfly layers per rotation is fixed per rotation, giving $n_\ell \cdot d/2$ angles per expert. This preserves sub-linear scaling while reducing training cost. For $d_{\text{model}}=256$ this yields 256 angles per input rotation.
    \item \textbf{Computation:} $O(n_\ell \cdot d)$ FLOPs per forward pass.
    \item \textbf{Expressivity:} Monarch matrices \cite{dao2022monarch} show that $O(\log d)$ butterfly layers approximate a broad class of orthogonal transformations with vanishing error as depth increases.
\end{itemize}

For inputs whose dimension is not a power of 2, we pad to the next power of two and strip the padding after each butterfly pass.

The entire butterfly layer is represented in algorithm-1.
\begin{algorithm}[t]
\caption{Butterfly Layer $\BF(\bm{\theta})$}
\label{alg:butterfly}
\begin{algorithmic}[1]

\Require Input $\mathbf{x} \in \mathbb{R}^{N \times d}$,
         angles $\bm{\theta} \in \mathbb{R}^{n_\ell \times d'/2}$
         where $d' = 2^{\lceil \log_2 d \rceil}$
\Ensure  Rotated output $\mathbf{x}' \in \mathbb{R}^{N \times d}$

\State \textbf{if} $d' > d$: zero-pad $\mathbf{x}$ to width $d'$

\For{$\ell = 1$ \textbf{to} $n_\ell$}
    \State $\mathbf{x}_{\text{even}} \leftarrow \mathbf{x}[:,\,0::2]$,\quad
           $\mathbf{x}_{\text{odd}}  \leftarrow \mathbf{x}[:,\,1::2]$
           \hfill $\triangleright$ split into paired channels
    \State $\mathbf{x}[:,\,0::2] \leftarrow
           \cos(\bm{\theta}_\ell)\odot\mathbf{x}_{\text{even}}
           - \sin(\bm{\theta}_\ell)\odot\mathbf{x}_{\text{odd}}$
    \State $\mathbf{x}[:,\,1::2] \leftarrow
           \sin(\bm{\theta}_\ell)\odot\mathbf{x}_{\text{even}}
           + \cos(\bm{\theta}_\ell)\odot\mathbf{x}_{\text{odd}}$
           \hfill $\triangleright$ block-diagonal Givens rotation
    \State $\mathbf{x} \leftarrow \text{PerfectShuffle}(\mathbf{x})$
           \hfill $\triangleright$ fixed permutation: interleave even/odd indices
\EndFor

\State \textbf{if} padded: strip last $d'-d$ columns
\State \Return $\mathbf{x}$

\end{algorithmic}
\end{algorithm}

\subsection{Quantized Substrate and Outlier Suppression}

\subsubsection{Ternary Quantization}
 $W_{\text{base}}$ is quantized to $\{-1, 0, +1\}$ (1.58 bits per weight) via AbsMean scaling\cite{ma2024era}:
\begin{equation}
    Q(W) = \gamma \cdot \text{round}\!\left(\frac{W}{\gamma + \epsilon}\right), \quad \gamma = \frac{1}{d_{\text{ff}} \cdot d_{\text{model}}} \sum_{ij} |W_{ij}|
\end{equation}
where $\text{round}(\cdot)$ clips to $\{-1, 0, +1\}$ and $\epsilon = 10^{-8}$ guards against division by zero.

\subsubsection{Straight-Through Estimator (STE)}
During backpropagation, $\frac{\partial Q(W)}{\partial W} = I$ is approximated, enabling gradient propagation despite the non-differentiability of the rounding operation \cite{bengio2013estimating}.

\subsubsection{Rotations for Outlier Suppression}
Activation distributions in transformer architectures exhibit extreme outliers. These are values that are 10 to 100$\times$ larger than the median
\cite{xiao2023smoothquant}.

Approaches like clipping remove high magnitude values which destroys information.Another approach for outlier suppresion is per channel scaling. Per channel scaling acts independently on each dimension and hence cannot distribute energy across dimensions\cite{ashkboos2024quarot}.

Rotations would be ideal in this scenario as quantization error is basis dependent. Orthogonal transformations preserve representational capacity while rotating activations into quantization friendly bases.

\subsubsection{Mechanism in ButterflyViT}
Every expert's input rotation  $\mathcal{B}(\theta_i)$ is jointly trained with $W_{\text{base}}$ under the total loss. As a result gradients $\frac{\partial \mathcal{L}}{\partial \theta_i}$ learn rotations that distribute activations with high magnitude across dimensions and align frequent activation patterns with low error regions of the ternary grid. This outlier suppression arises naturally without any explicit objective targeting quantization error. This happens because reducing ternary projection error directly reduces classification loss through the shared substrate.

\subsection{Training Objective and Orbital Initialization}

\subsubsection{Loss Function}
The total training loss is a collection of cross entropy loss for classification, load balance penalty to prevent expert routing collapse and a spatial smoothness regulariser. This is represented as:
\begin{equation}
\mathcal{L} = \mathcal{L}_{\text{CE}}(y, \hat{y})
            + \lambda_{\text{bal}}\, \mathcal{L}_{\text{bal}}
            + \lambda_{\text{sp}}\,  \mathcal{L}_{\text{sp}}
\end{equation}
with $\lambda_{\text{bal}} = 0.05$ and $\lambda_{\text{sp}} = 0.005$.

\paragraph{Load balance loss}
Uneven expert utilisation is penalised by squared expert load fractions. This is used in the Switch Transformer\cite{fedus2022switch} and is represented by:

\begin{equation}
\mathcal{L}_{\text{bal}} = N_E \sum_{i=1}^{N_E} f_i^2,
\qquad f_i = \frac{n_i}{B \cdot T \cdot k}
\end{equation}
where $n_i$ is the number of tokens routed to expert $i$, $B$ is batch size,
$T$ the number of patch tokens, and $k$ the top-$k$ value.
This penalises routing collapse quadratically and paves the way for uniform expert utilisation.

\paragraph{Spatial smoothness loss}
Standard MoE in ViT treats each token independently and ignores the spatial structure of the image patches. Patches are sometimes highly correlated in an image yet the unconstrained routing assigns them to different experts which introduces discontinuity. To fix this issue, a spatial smoothness regulariser penalises large differences in gate logits between temporally adjacent tokens. This is represented as:

\begin{equation}
\mathcal{L}_{\text{sp}} = \frac{1}{B(T-1)}
\left\| \mathbf{G}_{:,\,1:,\,:} - \mathbf{G}_{:,\,:-1,\,:} \right\|_F^2
\end{equation}
where $\mathbf{G} \in \mathbb{R}^{B \times T \times N_E}$ are the gate logits
reshaped into spatial order. 

This loss turns correlated patch tokens into a regularisation signal.

\subsubsection{Preventing Expert Collapse}
As all of the experts share the same ternary substrate $\mathbf{W}_{\text{base}}$, the rotation angles are the sole source of expert diversity. If all of the experts were initialised identically then all of the butterfly angles would get identical gradients and converge to the same solution, thereby negating the MoE capacity\cite{chi2022representation}.

This is prevented by initialising each expert's angles independently:

\begin{equation}
\bm{\theta}_i,\, \bm{\phi}_i \;\sim\; \mathcal{N}(0,\, 0.01^2)
\qquad \forall\; i \in \{1, \ldots, N_E\}
\end{equation}

The small variance is added to ensure that the rotations begin near the identity while the independent draws immediately break symmetry across experts. This is needed as $\mathbf{W}_{\text{base}}$ is shared, independent angle initialisation is the mechanism by which experts can develop unique views of the substrate from the first forward pass.

% In your document:

\begin{algorithm}[t]
\caption{ButterflyViT Forward Pass}
\label{alg:butterflyvit}
\begin{algorithmic}[1]

\Require Image $\mathbf{x}$, depth $L$, experts $\NE$, top-$k$
\Ensure  Logits $\mathbf{y}$, loss $\mathcal{L}$

\State $\mathbf{z} \leftarrow \text{PatchEmbed}(\mathbf{x}) + \mathbf{E}_{\text{pos}}$
\State $\mathcal{L}_{\text{bal}},\; \mathcal{L}_{\text{sp}} \leftarrow 0$

\For{$\ell = 1$ \textbf{to} $L$}
    \State $\mathbf{z} \leftarrow \mathbf{z} + \text{MHSA}(\text{LN}(\mathbf{z}))$
    \State $\mathbf{h} \leftarrow \text{LN}(\mathbf{z})$

    \State $\Wtern \leftarrow \text{TernaryQuant}(\mathbf{W}_{\text{base}})$
    \State $\mathbf{w},\,\mathbf{s} \leftarrow \text{TopK}(\mathbf{W}_{\text{gate}}\,\mathbf{h},\; k)$

    \State $\mathbf{out} \leftarrow \mathbf{0}$
    \For{$i = 1$ \textbf{to} $\NE$}
        \State $\mathbf{h}_i \leftarrow \mathbf{h}[\mathcal{M}_i]$
               \hspace{4.2em} $\triangleright$ tokens routed to expert $i$
        \State $\mathbf{h}_i \leftarrow \BF(\bm{\theta}_i,\, \mathbf{h}_i)$
               \hspace{2.85em} $\triangleright$ input butterfly rotation
        \State $\mathbf{u}_i \leftarrow \text{GELU}(\Wtern\, \mathbf{h}_i)$
               \hspace{1.5em} $\triangleright$ shared ternary up-projection
        \State $\mathbf{u}_i \leftarrow \BF(\bm{\phi}_i,\, \mathbf{u}_i)$
               \hspace{2.85em} $\triangleright$ output butterfly rotation
        \State $\mathbf{out}[\mathcal{M}_i] \mathrel{+}=
               w_i \cdot \mathbf{W}_{\downarrow}\,\mathbf{u}_i$
    \EndFor

    \State $\mathcal{L}_{\text{bal}} \mathrel{+}= \NE \sum_i f_i^2$
    \State $\mathcal{L}_{\text{sp}}  \mathrel{+}=
           \|\Delta_t\, \mathbf{G}\|_F^2$
    \State $\mathbf{z} \leftarrow \mathbf{z} + \mathbf{out}$
\EndFor

\State $\mathbf{y} \leftarrow \mathbf{W}_{\text{head}}\,\text{LN}(\mathbf{z})_{\text{cls}}$
\State $\mathcal{L} \leftarrow \mathcal{L}_{\text{CE}} +
       \lambda_{\text{bal}}\mathcal{L}_{\text{bal}} +
       \lambda_{\text{sp}}\mathcal{L}_{\text{sp}}$
\State \Return $\mathbf{y},\;\mathcal{L}$

\end{algorithmic}
\end{algorithm}

\subsection{Algorithm}
The entire architecture of ButterflyViT is represented in Algorithm-2.

\subsection{Theoretical Properties}
Two theoretical properties define the memory and compression lower bound of ButterflyViT. The exact memory scaling of the orbital parameterization is derived along with the asymptotic compression ratio which is established relative to standard ViT-MoE architectures.

\begin{proposition}[Memory Scaling]
For $N_E$ experts with dimensions $d_{\text{model}}$, $d_{\text{ff}}$, and $n_\ell$ butterfly layers, ButterflyViT expert memory is:
\begin{equation}
M_{\text{ButterflyViT}} = \frac{1.58}{8}\, d_{\text{ff}}\, d_{\text{model}} + N_E \cdot 4\, n_\ell \left(\frac{d_{\text{model}}}{2} + \frac{d_{\text{ff}}}{2}\right) \cdot 2
\end{equation}
\end{proposition}

\paragraph{Justification.}
ButterflyViT replaces independent expert FFNs with a single shared ternary matrix $\mathbf{W}_{\text{base}} \in \{-\gamma, 0, +\gamma\}^{d_{\text{ff}} \times d_{\text{model}}}$ plus per-expert learnable rotation angles. At 1.58 bits per ternary weight the substrate occupies $\frac{1.58}{8}\,d_{\text{ff}}\,d_{\text{model}}$ bytes regardless of $N_E$. Each expert $i$ is fully described by two modules: an input rotation $\bm{\theta}^{(i)} \in \mathbb{R}^{n_\ell \times d_{\text{model}}/2}$ and an output rotation $\bm{\phi}^{(i)} \in \mathbb{R}^{n_\ell \times d_{\text{ff}}/2}$, where each layer stores one Givens angle per adjacent channel pair. Both parameter tensors are stored in FP16, contributing $n_\ell\,\frac{d_{\text{model}}}{2} \cdot 2 + n_\ell\,\frac{d_{\text{ff}}}{2} \cdot 2$ bytes per expert. Multiplying by $N_E$ and adding the fixed substrate gives the stated result.

\begin{proposition}[Compression Lower Bound]
The compression ratio of ButterflyViT over Standard MoE grows monotonically with $N_E$ and is bounded below by:
\begin{align}
\frac{M_{\text{Standard MoE}}}{M_{\text{ButterflyViT}}}
&\;\geq\;
\lim_{N_E \to \infty} \frac{M_{\text{Standard MoE}}}{M_{\text{ButterflyViT}}} \notag \\
&= \lim_{N_E \to \infty}
\frac{N_E \cdot 4\, d_{\text{model}}\, d_{\text{ff}}}
{\dfrac{1.58}{8}\,d_{\text{ff}}\,d_{\text{model}}
+ N_E \cdot 2\,n_\ell\!\left(d_{\text{model}} + d_{\text{ff}}\right) \cdot 2} \notag \\
&= \frac{4\, d_{\text{model}}\, d_{\text{ff}}}
{2\, n_\ell\!\left(d_{\text{model}} + d_{\text{ff}}\right) \cdot 2} \notag \\
&\hspace{3em}\left(\text{} \tfrac{1.58}{8}\,d_{\text{ff}}\,d_{\text{model}} \text{ is fixed, vanishes relative to } N_E \to \infty\right) \notag \\
&= \frac{d_{\text{model}}\, d_{\text{ff}}}{n_\ell\!\left(d_{\text{model}} + d_{\text{ff}}\right)}
\end{align}
For $d_{\text{model}} = 256$, $d_{\text{ff}} = 1024$, $n_\ell = 2$:
\begin{align*}
&= \frac{256 \times 1024}{2 \times (256 + 1024)} \\[4pt]
&= \frac{256 \times 1024}{2 \times 1280} \\[4pt]
&= \frac{262{,}144}{2{,}560} \\[4pt]
&\approx 409\times
\end{align*}
At $N_E$=64, the substrate quantization contributes to additional compression yielding observed 354 $\times$ compression. As $N_E$ increases, the compression ratio increases.
\end{proposition}

\section{Results}
ButterflyViT was trained and evaluated on the CIFAR-100 dataset(50000 images for training, 10000 for evaluation). This dataset was chosen as expert specialization and memory compute trade offs can be shown. Our model uses
$d_{\text{model}} = 256$,
$d_{\text{ff}} = 1024$,
and $N_E = 8$ experts with top-$k=2$.
\subsection{Training Curves}
ButterflyViT is compared against Standard ViT-MoE based architecture and with dense FFN baselines. All of the architectures are trained for 50 epochs, and a cosine based decay with a linear warmup is implemented. 
\begin{figure}[h]
    \centering
    \includegraphics[width=0.5\textwidth]{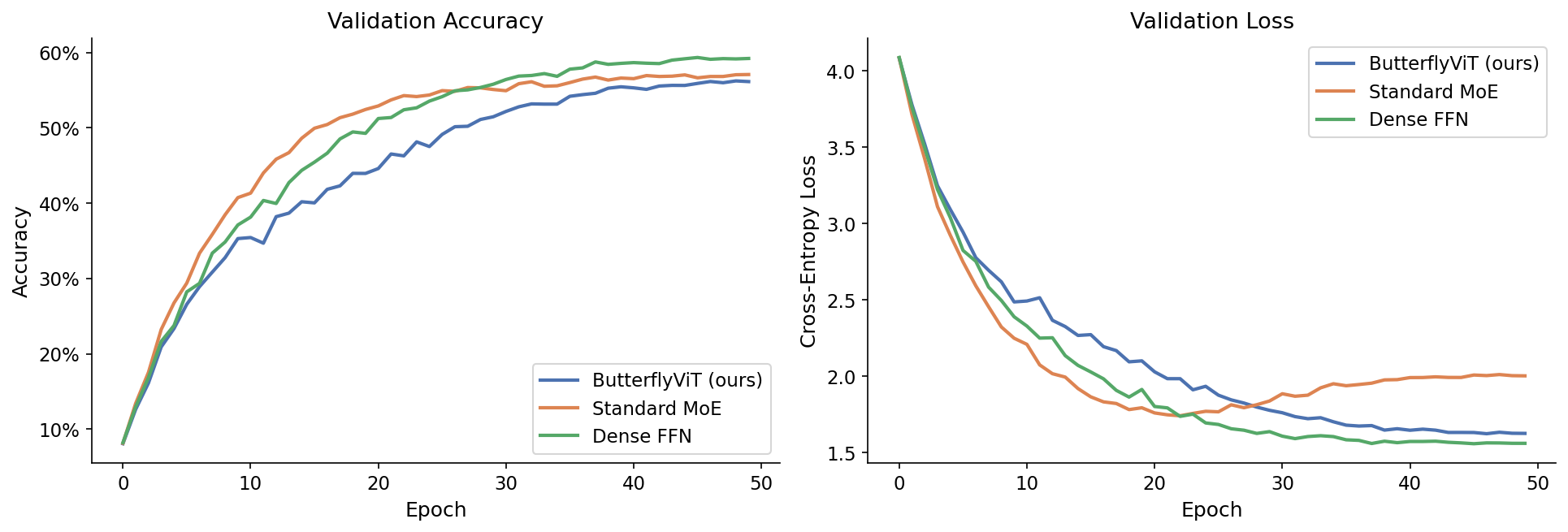} % Replace "example-image" with your image file name
    \caption{Training Curves and Validation loss visualised.}
    \label{fig:A}
\end{figure}

\begin{table}[h]
\centering
\caption{Comparison of Model Performance and Resource Usage. Here the parameters are measured in millions, memory is measured in MB}
\label{tab:model_comparison}
\begin{tabular}{lcccc}
\toprule
\textbf{Model} & \textbf{Val Acc} & \textbf{Params} & \textbf{Mem} & \textbf{Compression} \\
\midrule
ButterflyViT & 56.24\% & 5.66 & 0.64 & 181$\times$ \\
Standard MoE & 57.09\% & 31.35 & 117.44 & 1$\times$ \\
Dense FFN & 59.35\% & 5.58 & 117.44 & 1$\times$ \\
\bottomrule
\end{tabular}
\end{table}

Table-1 shows that the expert memory in ButterflyViT is compressed by 181 times while maintaining similar accuracy. ButterflyViT reparameterizes experts as structural transformations on a single substrate this eliminates the need to save separate expert matrices.

From figure 3 and table-1, we see that Dense FFN has a slightly higher accuracy than both of the ViT architectures. This is because the dense baseline has the full $d_{\text{model}} \times d_{\text{ff}}$ matrices whereas ButterflyViT's matrices are ternary quantized and experts share it. Datasets like CIFAR-100, might not show the full effectiveness of the MoE architecture as dense FFN has more effective capacity per parameter.

\subsection{Memory Scaling Analysis}

\begin{figure}[h]
    \centering
    \includegraphics[width=0.5\textwidth]{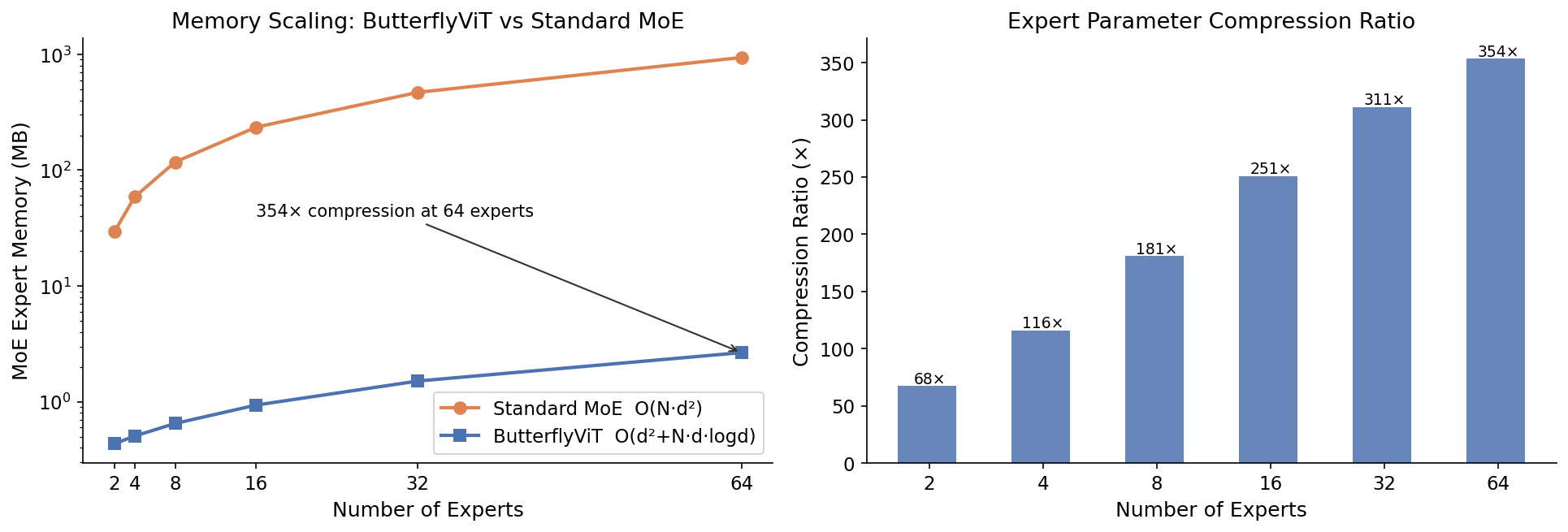} % Replace "example-image" with your image file name
    \caption{(Left) Number of experts is compared to the MoE expert memory. (Right) Expert Parameter Compression Ratio }
    \label{fig:A}
\end{figure}

Figure-4 compares memory scaling with StandardMoE and ButterflyViT. From the graph it is evident that ButterflyViT scales better and keeps the MoE expert memory relatively small. This is due to ButterflyViT's sublinear scaling which is not the case for StandardMoE.

\begin{table}[h]
\centering
\caption{Comparison of Expert Memory Usage}
\label{tab:expert_memory}
\begin{tabular}{lrrr}
\toprule
\textbf{Experts} & \textbf{StandardMoE} & \textbf{ButterflyViT} & \textbf{Compression} \\
\midrule
2  & 29.36  & 0.434  & 68$\times$ \\
4  & 58.72  & 0.505  & 116$\times$ \\
8  & 117.44 & 0.649  & 181$\times$ \\
16 & 234.88 & 0.935  & 251$\times$ \\
32 & 469.76 & 1.509  & 311$\times$ \\
64 & 939.52 & 2.656  & 354$\times$ \\
\bottomrule
\end{tabular}
\end{table}

Table-2 shows the comparison of expert memory usage. From the table, we see that as the number of experts increase the compression ratio increases. When the number of experts is 64, the expert memory for StandardMoE is 939.52 compare to 2.65 for ButterflyViT achieving a 354$\times$ compression.

\subsection{Expert Similarity}

Experts maintain distinct behaviour even though they operate a shared quantized substrate. Figure 5 shows the pairwise cosine similarity for both ButterflyViT and StandardMoE. ButterflyViT has an off diagonal mean of 0.29 compared to StandardMoE's 0.10. The higher similarity in ButterflyViT shows that the shared substrate introduced an expert manifold rather than independent expert specialisation. StandardMoE's outputs are very orthogonal showing that all of the experts learned completely independently with no parameter sharing. This is shown in Figure 5.

\begin{figure*}[t]
    \centering
    \begin{subfigure}[b]{0.46\textwidth}
        \includegraphics[width=\textwidth]{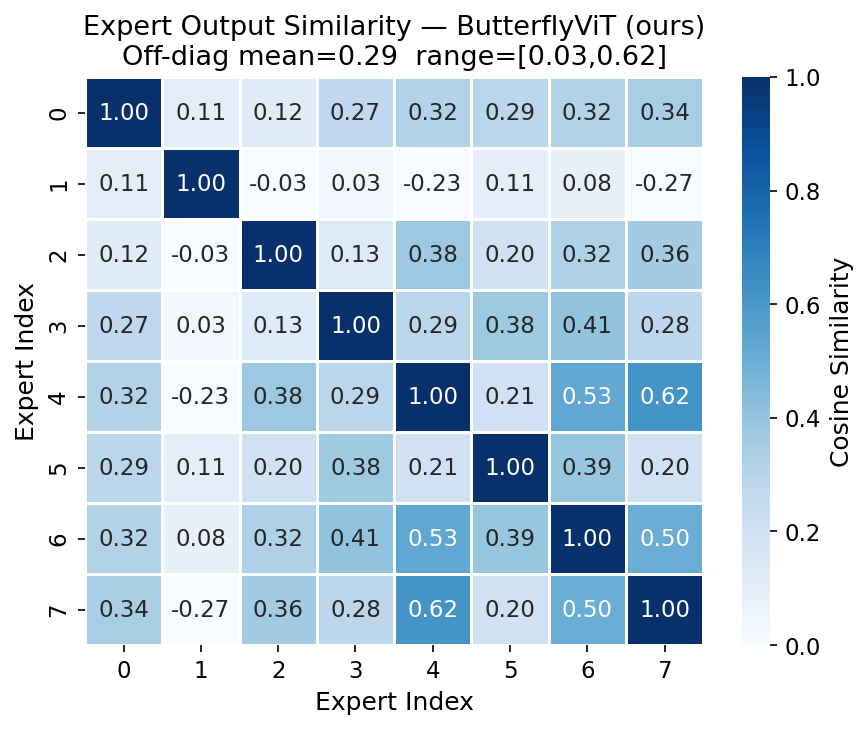} % Replace "image1" with your first image filename
        \caption{ButterflyViT-Cosine Similarity Matrix}
        \label{fig:image1}
   \end{subfigure}
    \hfill
    \begin{subfigure}[b]{0.46\textwidth}
        \includegraphics[width=\textwidth]{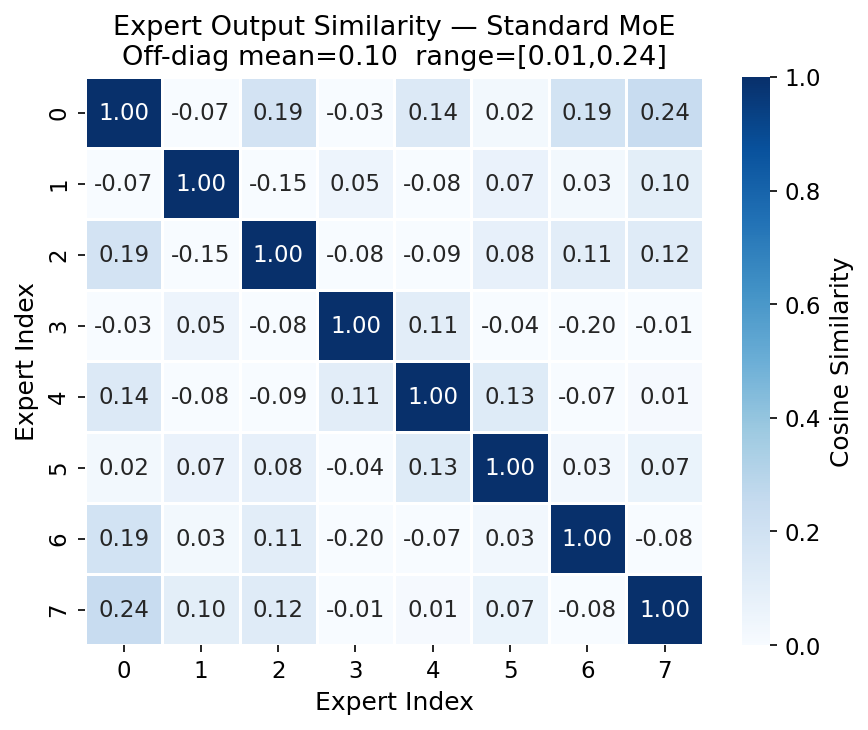} % Replace "image2" with your second image filename
        \caption{StandardMoE Cosine Similarity Matrix}
        \label{fig:image2}
    \end{subfigure}
    \caption{Cosine Similarity showing the similarity score across 8 experts.}
    \label{fig:side_by_side}
\end{figure*}

\subsection{Real World Deployment and Energy Analysis}
\begin{figure}[H]
    \centering
    \includegraphics[width=0.5\textwidth]{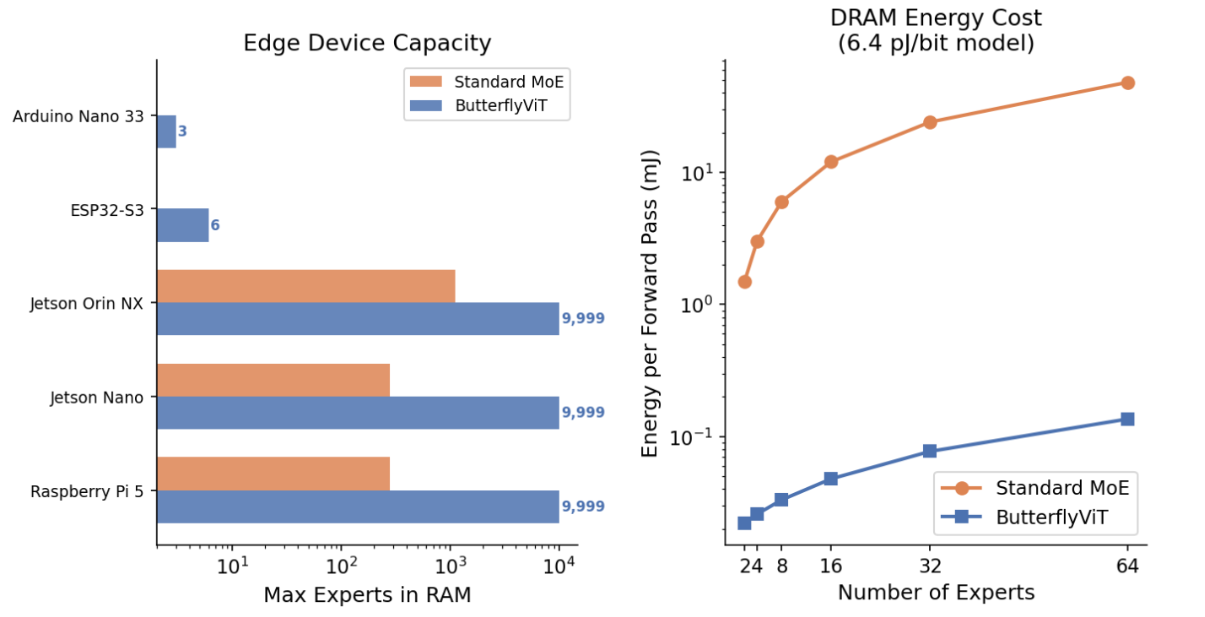} % Replace "example-image" with your image file name
    \caption{Edge Deployment and Energy Analysis}
    \label{fig:A}
\end{figure}
ButterflyViT touches 9999 experts on edge devices like Jetson Nano and Raspberry Pi whereas StandardMoE fits a few hundred. On microcontrollers like ESP32-S3, Arduino Nano 33, ButterflyViT fits 3-6 experts whereas StandardMoE can fit none.) ButterflyViT remains below 0.2mJ in DRAM energy cost per forward pass versus 90 mJ for StandardMoE. This consistently results in more than 98\% of energy savings for ButterflyViT which is shown in the figure below.

\begin{figure}[H]
    \centering
    \includegraphics[width=0.4\textwidth]{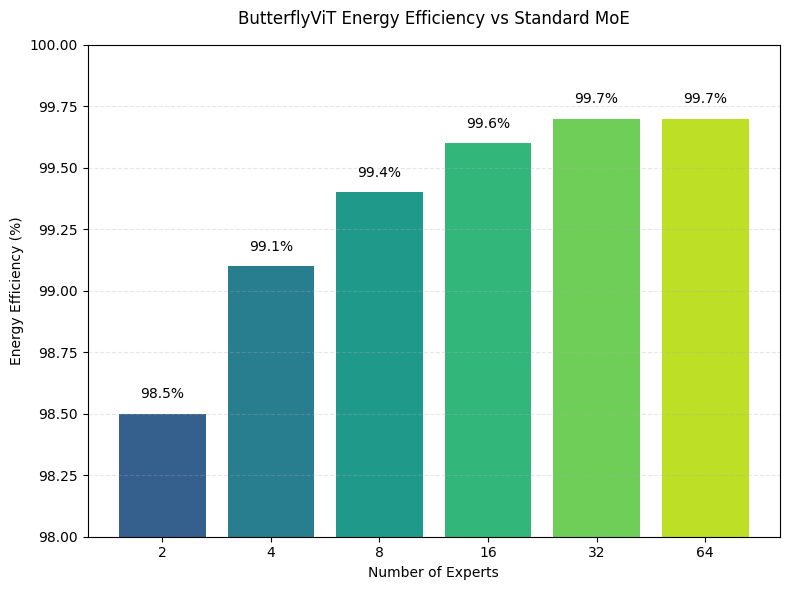} % Replace "example-image" with your image file name
    \caption{Energy efficency analysis}
    \label{fig:A}
\end{figure}
\section{Conclusion}

ButterflyViT achieves sub linear memory scaling in ViT based MoE architectures through parameterization of experts as orbital transformations of a shared ternary substrate. This methodology yields more than 350 $\times$ compression at 64 experts while maintaining accuracy.
Rather than treating experts as independent parameter sets, a geometrical structure is implemented that avoids the linear memory barrier for sparse architectures. This positions group-orbit representations as a mechanism for extreme compression without expert collapse.
ButterflyViT was evaluated and tested on Kaggle's T4x2 GPUs with ButterflyViT being 3 $\times$ slower than StandardMoe. This was later fixed using a custom Triton kernel which showed ButterflyViT achieving near same inference speed as StandardMoE

Future work will include large scale datasets, bigger parameters, interpreting rotations and exploring this idea in other architectures.

To our knowledge, no prior ViT-MoE treats expert parameter compression as its main objective. This work establishes the first benchmark on this axis.

% References (use BibTeX or manual entries)
\bibliographystyle{plainnat}
\bibliography{references}

\end{document}